\documentclass[conference]{IEEEtran}
\IEEEoverridecommandlockouts
\usepackage{courier}
\usepackage{booktabs}
\usepackage{url}
\usepackage{multirow}
\usepackage{threeparttable}
\usepackage{cite}
\usepackage{amsmath,amssymb,amsfonts}
\usepackage{algorithmic}
\usepackage{graphicx}
\usepackage{textcomp}
\usepackage{xcolor}
\usepackage[none]{hyphenat}
\usepackage[font=footnotesize,subrefformat=parens,labelformat=parens]{subfig}

\usepackage{tikz}

\newcommand\LADot{%
  \mathord{%
    \mspace{1mu}%
    \text{\ladot}%
    \mspace{1mu}%
  }%
}
\newcommand\ladot{%
    \tikz[line cap=round,x=1ex,y=1ex,line width=0.3pt]
    {\draw (0,1) |- (1,0) -- (1,-1);}%
}

\newcommand\RADot{%
  \mathord{%
    \mspace{1mu}%
    \text{\radot}%
    \mspace{1mu}%
  }%
}
\newcommand\radot{%
    \tikz[line cap=round,x=1ex,y=1ex,line width=0.3pt]
    {\draw (1, 2) -- (1,1) -| (0,0);}%
}

\def\BibTeX{{\rm B\kern-.05em{\sc i\kern-.025em b}\kern-.08em
    T\kern-.1667em\lower.7ex\hbox{E}\kern-.125emX}}

\begin{document}

\title{PubLayNet: largest dataset ever for document layout analysis}

\author{\IEEEauthorblockN{Xu Zhong}
\IEEEauthorblockA{\textit{IBM Research Australia} \\
60 City Road, Southbank \\
 VIC 3006, Australia \\
peter.zhong@au1.ibm.com}
\and
\IEEEauthorblockN{Jianbin Tang}
\IEEEauthorblockA{\textit{IBM Research Australia} \\
60 City Road, Southbank\\
 VIC 3006, Australia \\
jbtang@au1.ibm.com}
\and
\IEEEauthorblockN{Antonio Jimeno Yepes}
\IEEEauthorblockA{\textit{IBM Research Australia} \\
60 City Road, Southbank \\
VIC 3006, Australia \\
antonio.jimeno@au1.ibm.com}
}

\maketitle

\begin{abstract}
Recognizing the layout of unstructured digital documents is an important step when parsing the documents into structured machine-readable format for downstream applications. Deep neural networks that are developed for computer vision have been proven to be an effective method to analyze layout of document images. However, document layout datasets that are currently publicly available are several magnitudes smaller than established computing vision datasets. Models have to be trained by transfer learning from a base model that is pre-trained on a traditional computer vision dataset. In this paper, we develop the PubLayNet dataset for document layout analysis by automatically matching the XML representations and the content of over~1~million PDF articles that are publicly available on PubMed Central\textsuperscript{\texttrademark}. The size of the dataset is comparable to established computer vision datasets, containing over 360 thousand document images, where typical document layout elements are annotated. The experiments demonstrate that deep neural networks trained on PubLayNet accurately recognize the layout of scientific articles. The pre-trained models are also a more effective base mode for transfer learning on a different document domain. We release the dataset (\url{https://github.com/ibm-aur-nlp/PubLayNet}) to support development and evaluation of more advanced models for document layout analysis.
\end{abstract}


\section{Introduction}

Documents in Portable Document Format (PDF) are ubiquitous with around 2.5 trillion documents available in this format~\cite{staar2018ccs}.
While these documents are convenient for human consumption, automatic processing of these documents is difficult since understanding the document layout and extracting information using this format is complicated.

Geometric layout analysis techniques based on an image representation of the document combined with optical character recognition (OCR) methods~\cite{cattoni1998geometric,breuel2002two,breuel2003high} were firstly used to understand these documents.
More recently, image analytics methods based on deep learning are becoming available~\cite{he2017mask} and are used to train document layout understanding pipelines~\cite{staar2018ccs,schreiber2017deepdesrt}.

Machine learning methods require training data to become successful.
In addition, there is a large variety of document templates, which makes this task even more challenging since this increases the number of documents that need to be manually annotated.
On the other hand, manual annotation is a slow and expensive process, which is a stepping curve when willing to use these techniques in new domains.

In this work, we propose a method to automatically annotate the document layout of over~1~million PubMed Central\textsuperscript{\texttrademark} PDF articles and generate a high-quality document layout dataset called PubLayNet. The dataset contains over 360k page samples and covers typical document layout elements such as text, title, list, figure, and table. Then, we evaluate deep object detection neural networks on the PubLayNet dataset and the performance of fine tuning the networks on existing small manually annotated corpora. We show that the automatically annotated dataset is suitable to train a model to recognize the layout of scientific articles, and the model pre-trained on the dataset can be a more effective base in transfer learning.

\section{Related work}

Existing datasets for document layout analysis rely on manual annotation.
Some of these datasets are used in document processing challenges.
Examples of these efforts are available in several ICDAR challenges~\cite{antonacopoulos2009realistic}, which cover as well complex layouts~\cite{clausner2015enp,clausner2017icdar2017}.
The US NIH National Library of Medicine has provided the Medical Article Records Groundtruth (MARG)\footnote{\url{https://ceb.nlm.nih.gov/inactive-communications-engineering-branch-projects/medical-article-records-groundtruth-marg/}}, which are obtained from scanned article pages.

In addition to document layout, further understanding of the document content has been studied in the evaluation of table detection methods, e.g.~\cite{shahab2010open,fang2012dataset}.
Examples include table detection from document images using heuristics~\cite{yildiz2005pdf2table}, vertical arrangement of text blocks~\cite{tran2015table} and deep learning methods~\cite{he2017multi,hao2016table,gilani2017table,kavasidis2018saliency}.

Overall, the datasets are of limited size, just several hundred pages, which is mostly due to the need for manual annotation.
In the next section, we describe how multiple versions of the same document from PubMed Central\textsuperscript{\texttrademark} are used to automatically generate document layout annotations for over 1 million documents.

\section{Automatic annotation of document layout}

\begin{figure*}[!htb]
  \centering
  \begin{minipage}[b]{.25\linewidth}%
  \subfloat[PDF representation.]{\includegraphics[width=\linewidth]{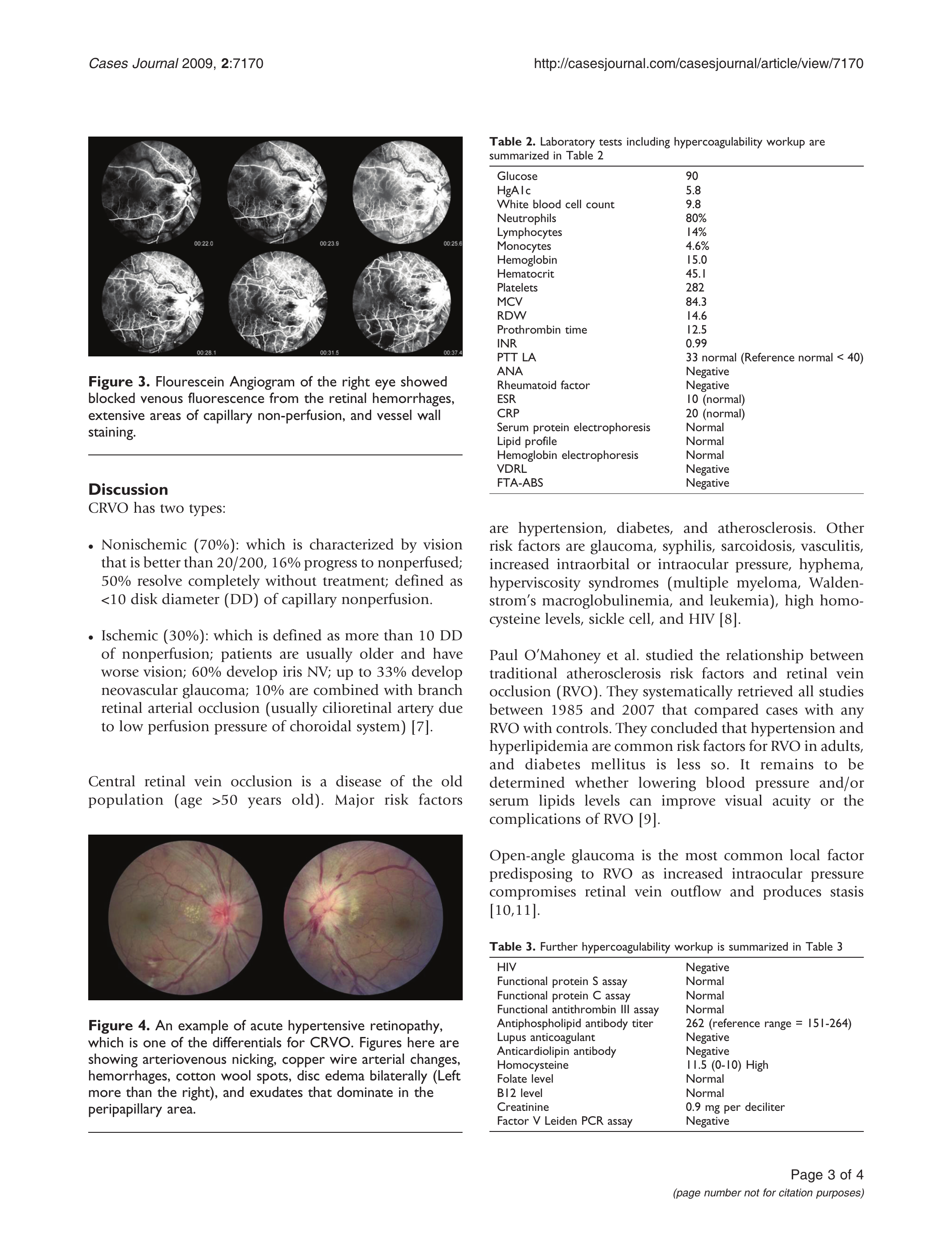}\label{fig:pdf_page}}%
  \end{minipage}%
  \begin{minipage}[b]{.25\linewidth}%
  \centering%
  \subfloat[XML representation.]{\includegraphics[width=0.83\linewidth]{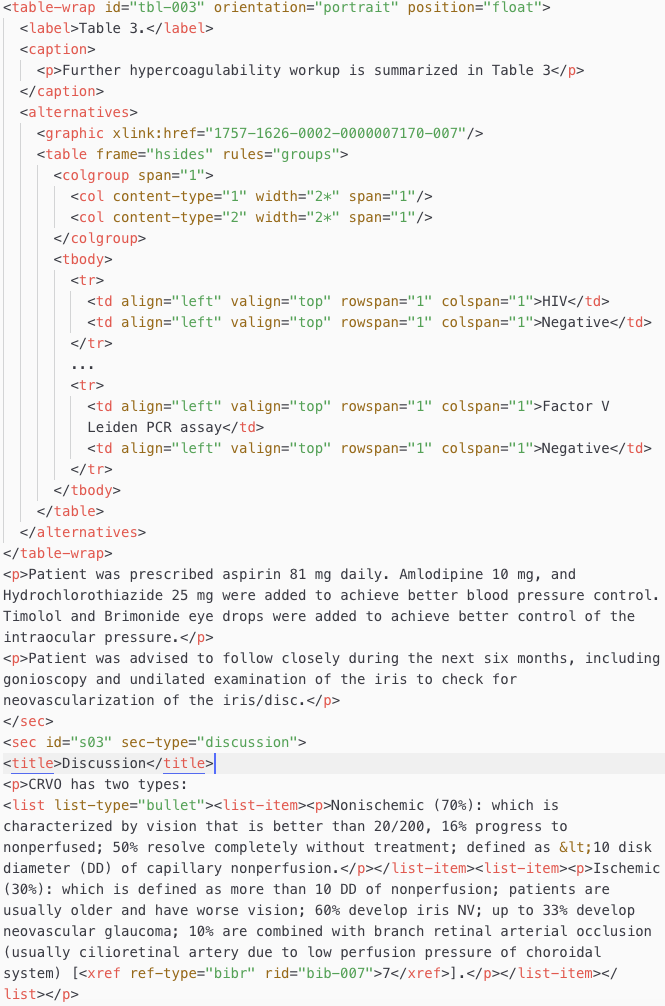}\label{fig:xml}}%
  \end{minipage}%
  \begin{minipage}[b]{.25\linewidth}%
  \subfloat[\texttt{PDFMiner} output on (a).]{\includegraphics[width=\linewidth]{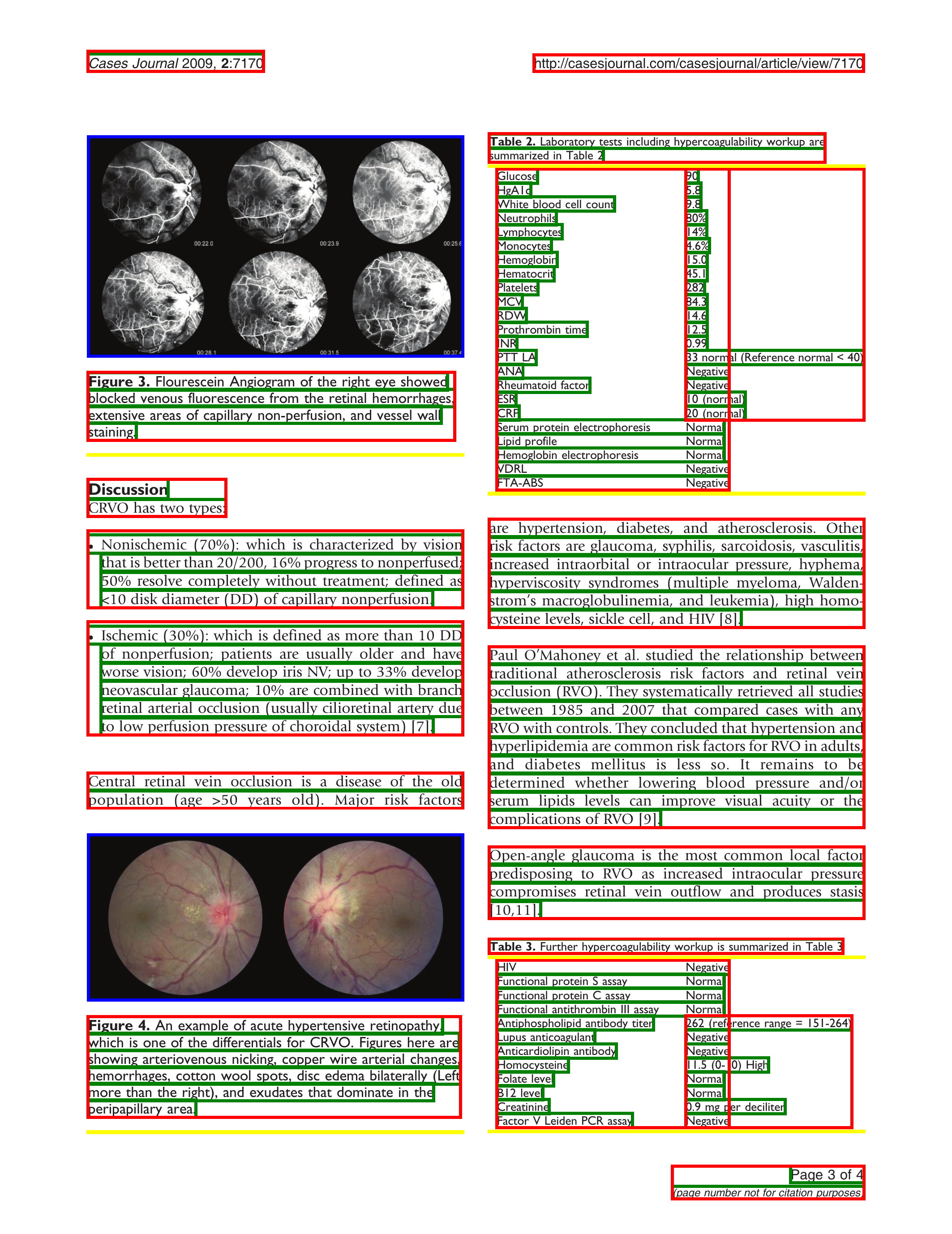}\label{fig:pdf_data}}
  \end{minipage}%
  \begin{minipage}[b]{.25\linewidth}%
  \subfloat[Annotations generated by matching (b) and (c).]{\includegraphics[width=\linewidth]{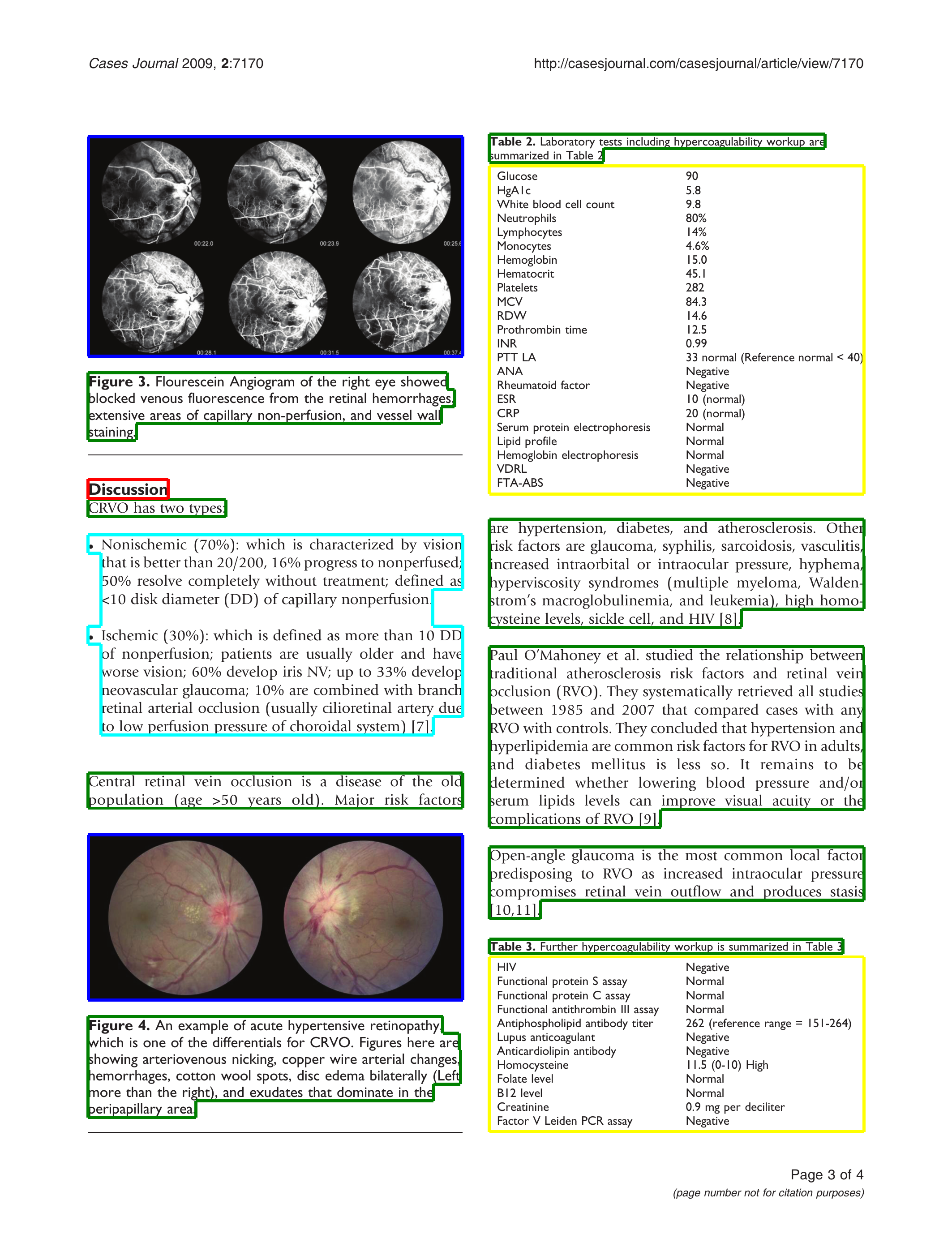}\label{fig:annotation}}
  \end{minipage}%
  \caption{Parsing PDF page (a) using \texttt{PDFMiner} (c) and matching the layout with the XML representation (b) to generate annotation of page layout (d). The color scheme in (c) is red: textbox; green: textline; blue: image; yellow: geometric shape. The color scheme in (d) is red: title; green: text; blue: figure; yellow: table; cyan: list.}
\end{figure*}

To overcome the lack of training data, we used a large document collection from PubMed Central\textsuperscript{\texttrademark} Open Access (PMCOA), provided under the Creative Commons license. The articles in PMCOA are provided in both PDF format (Fig.~\ref{fig:pdf_page}) and in an XML format (Fig.~\ref{fig:xml}). The XML version of the PMCOA documents is a structured representation of its content and all XML documents follow the schema provided by the NLM for journals\footnote{https://dtd.nlm.nih.gov}. Since the content in the PDF version of the articles and their XML representation contain similar format, we have identified a method to use these two representations of the same article to identify document layout components. In this work, a total of 1,162,856 articles that have a complete XML representation were downloaded from \url{ftp.ncbi.nlm.nih.gov/pub/pmc} on 3 October 2018, and automatically annotated with the method described in the following sections.

\subsection{Layout categories}

The structured XML representation of the articles in the PMCOA dataset contains many different categories of nodes, which are difficult, even for humans, to distinguish based only on document images. We aggregated the categories of the nodes in the XML into the document layout categories shown in Table~\ref{tab:category}, based on the following considerations:
\begin{itemize}
  \item The differences between the categories are distinctive and intuitive for a visual model to capture learnable patterns.
  \item The categories are commonly found in documents in various domains.
  \item The categories cover most elements that are important for downstream studies, such as text classification, entity recognition, figure/table understanding, etc.
\end{itemize}

\begin{table}[!htb]
  \caption{Categories of document layout included in PubLayNet.}
  \label{tab:category}
  \centering
  \begin{threeparttable}
  \begin{tabular}{p{2cm}p{5.5cm}}
    \toprule
    Document layout category & XML category \\
    \midrule
    Text & author, author affiliation; paper information; copyright information; abstract; paragraph in main text, footnote, and appendix; figure \& table caption; table footnote\\
    Title & article title, standalone (sub)section title\tnote{a}, standalone figure \& table label\tnote{b} \\
    List & list\tnote{c} \\
    Table & main body of table \\
    Figure & main body of figure\tnote{d} \\
    \bottomrule
  \end{tabular}
  \begin{tablenotes}\footnotesize
  \item[a] When a section title is inline with the leading text of the section, it is labeled as part of the text, but not as a title.
  \item[b] When a figure/table label is inline with the caption of the figure/table, it is labeled as part of the text, but not as a title.
  \item[c] Nested lists (i.e., child list under an item of parent list) are annotated as a single object. Child lists in nested lists are not annotated.
  \item[d] When sub-figures exist, the whole figure panel is annotated as a single object. Sub-figures are not annotated individually.
  \end{tablenotes}
  \end{threeparttable}
\end{table}

\subsection{Annotation algorithms}

Our annotation algorithm matches PDF elements (see Section~\ref{sec:pdf}) to the XML nodes. Then, the bounding box and segmentation of the PDF elements are calculated. The XML nodes are used to decide the category label for each bounding box and segmentation. Finally, a quality control metric is defined to control the noise of the annotations at an extremely low level.

\subsubsection{PMCOA XML pre-processing and parsing}

Some of the nodes in the XML tree are not considered for matching, such as \emph{tex-math}, \emph{edition}, \emph{institution-id}, and \emph{disp-formula}. These nodes are removed, as the content of these nodes may interfere with the matching of other nodes. The placement of list, table and figure nodes in the XML schema is not consistent across the articles. We standardized the XML tree by moving list, table, and figure nodes into the floats-group branch. Then, the nodes in the XML tree are split into five groups:
\begin{itemize}
  \item Sorted: including paper title, abstract, keywords, section titles, and text in main text. The order of sorted XML nodes matches the reading order in the PDF document.
  \item Unsorted: including copyright statement, license, authors, affiliations, acknowledgments, and abbreviations. The order of unsorted XML nodes may not match the reading order in the PDF document.
  \item Figures: including caption label (e.g., `Fig.~1'), caption text, and figure body
  \item Tables: including caption label (e.g. Table I), caption text, footnotes, and table body
  \item Lists: including lists.
\end{itemize}

\subsubsection{PMCOA PDF parsing}
\label{sec:pdf}
Fig.~\ref{fig:pdf_data} illustrates an example of the layout of a PDF page parsed using the \texttt{PDFMiner}\footnote{\url{https://github.com/euske/pdfminer}} package, where three layout types are extracted:
\begin{itemize}
  \item textbox (red): block of text, consisting of textlines (blue). Each textbox has three attributes: the text in the textbox, the bounding box of the textbox, and textlines in the textbox. Each textline has two attributes: the text in the textline and the bounding box of the textline.
  \item image (green): consisting of images. Each image is associated with a bounding box.
  \item geometric shape (yellow): consisting of lines, curves, and rectangles. Each geometric shape is associated with a bounding box.
\end{itemize}

\subsubsection{String pre-processing}

The strings from XML and PDF are Unicode strings. The Unicode standard defines various normalization forms of a Unicode string, based on the definition of canonical equivalence and compatibility equivalence. In Unicode, several characters can be expressed in various ways. To make the matching between XML and PDF more robust, the strings are normalized to the KD normal form\footnote{\url{http://unicode.org/reports/tr15/}} (i.e., replacing all compatibility characters with their equivalents).

\subsubsection{PDF-XML matching algorithms}

There are frequent minor discrepancies between the content of PDF parsed by \texttt{PDFMiner} and the text of XML nodes. Thus fuzzy string matching is adopted to tolerate minor discrepancies. We use the fuzzysearch\footnote{\url{https://github.com/taleinat/fuzzysearch}} package to search for the closest match to a target string in a source string, where string distance is measured by the Levenshtein distance~\cite{levenshtein1966binary}. The maximum distance allowed for a match ($d_{max}$) is adaptive to the length of the target string ($l_{target}$) as,
\begin{equation*}
  d_{max} = \left\{
            \begin{array}{ll}
            0.2 * l_{target} & \text{if}\ l_{target} \le 20 \\
            0.15 * l_{target} & \text{if}\ 20 < l_{target} \le 40 \\
            0.1 * l_{target} & \text{if}\ l_{target} > 40 \\
            \end{array}
            \right.
\end{equation*}

As one textbox may cover multiple XML nodes, we sequentially search the textlines of a textbox in the text of a XML node. If the textline of the textbox cannot be found in the text of the XML node, we skip to and search the next textbox. If the end of the XML node is reached, but the textline is not the end of the textbox, the textbox is divided into two textboxes at the current textline. Then the former textbox is appended to the list of matched textboxes of the XML node. When all the content of the XML node is covered by matching textlines, we start searching in the next XML node. This matching procedure is applied to all the text XML nodes, including the `Sorted', `Unsorted', and `Lists' groups; the captions in the `Tables' and `Figures' groups; and the footnotes in the `Table' group.

Depending on the template of specific journals, section/subsection titles may be inline with the first paragraph in the section. A title is treated as inline titles if the last line of the title does not cover a whole textline. Inline section titles are annotated as part of the text, rather than individual instances of titles. The same principle is also applied to the caption labels of figures and tables.

\begin{figure}[!htb]
  \centering
  \includegraphics[width=.6\linewidth]{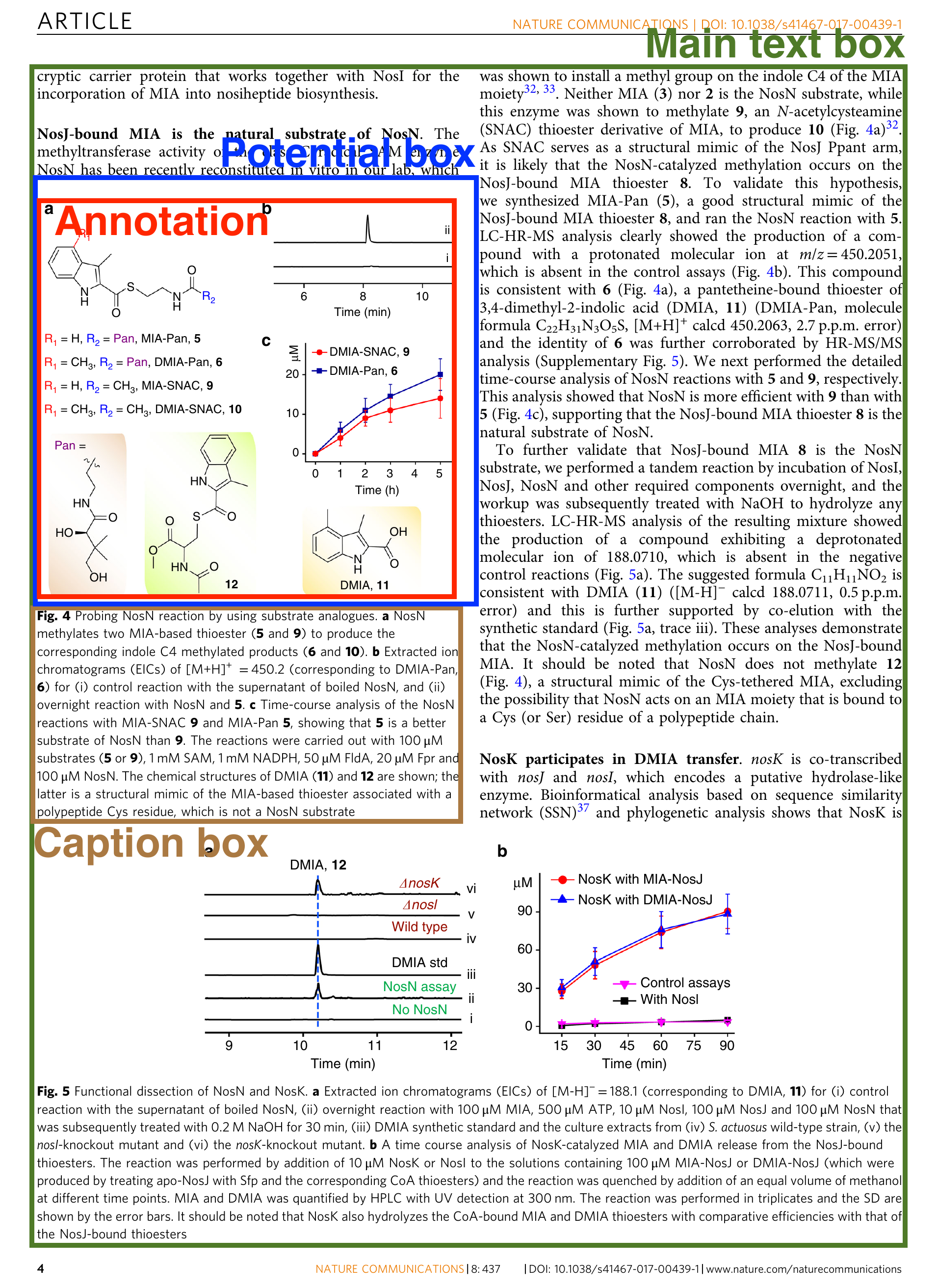}
  \caption{Annotation process for an example figure. The main text box and the potential box are determined from the annotations of the caption of the figure and surrounding text elements. The final annotation is made as the smallest box that encloses all the textboxes, images, and geometric shapes within the potential box.}
  \label{fig:box}
\end{figure}

After all the text XML nodes are processed, the margin between annotated text elements in the PDF page is utilized to annotate the body of figures and tables. Fig.~\ref{fig:box} illustrates an example of the annotation process for a figure body. First, the bounding box of the main text of the article (green box) is obtained as the smallest bounding box that encloses all the annotated text elements in the article. Then the potential box (blue box) for the figure is calculated as the the largest box that can be fit in the margin between the top of the caption box (brown box) and the annotated text elements above the caption. The last step is to annotate the figure body with the smallest box (red box) that encloses all the textboxes, images, and geometric shapes within the potential box. Table bodies are annotated using the same principle, where it is assumed that table bodies are always below the caption of the tables.

\subsubsection{Generation of instance segmentation}

\begin{figure}[!htb]
  \centering
  \includegraphics[width=.8\linewidth]{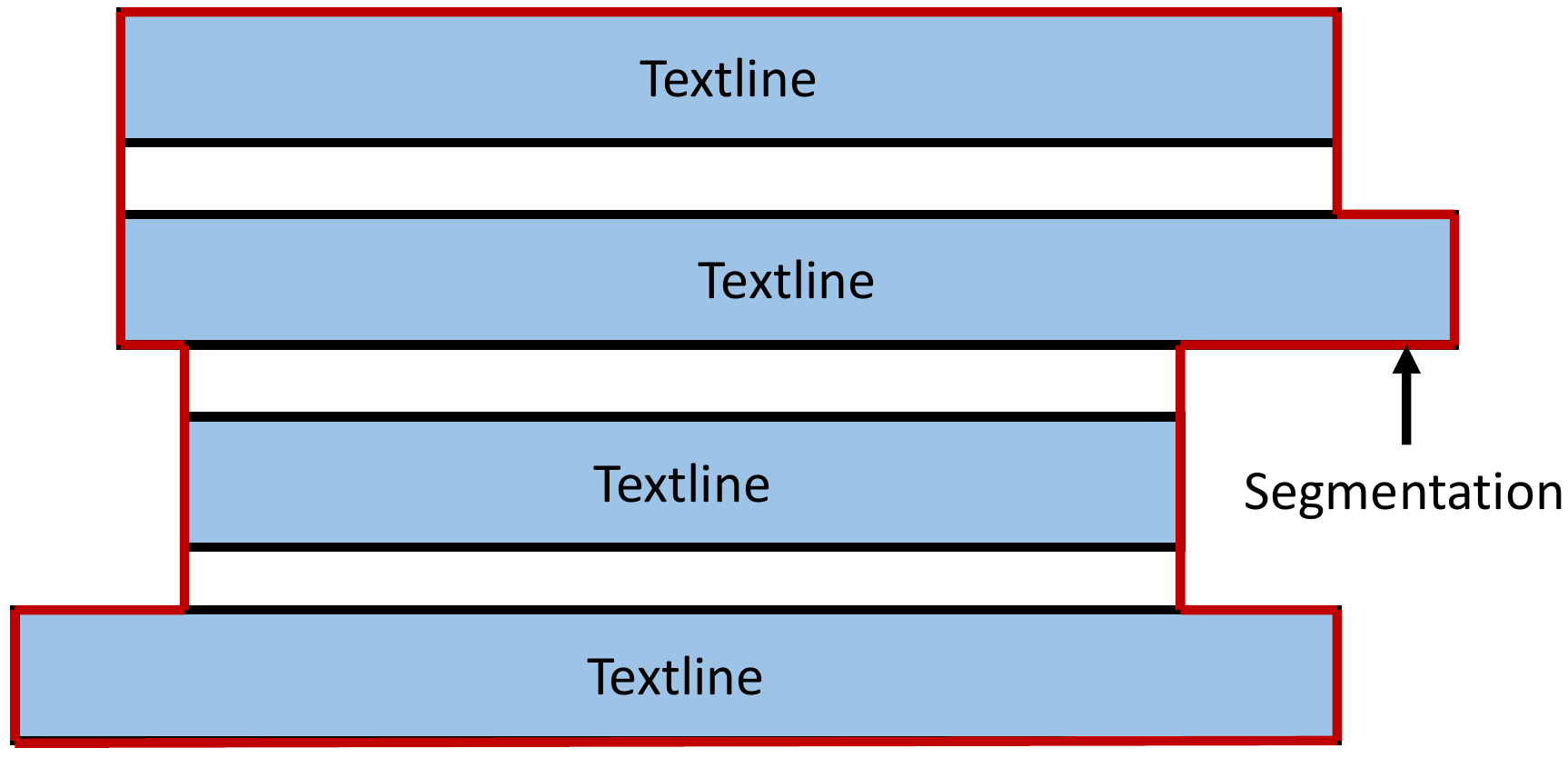}
  \caption{Example of generating layout segmentation based on textlines. The segmentation is a regular polygon, consisting of only horizontal and vertical edges. The shape of the segmentation is decided by the position of adjacent textlines.}
  \label{fig:seg}
\end{figure}

For text, title, and list instances, we automatically generate segmentations from the texelines of the PDF elements, which allows us to train the Mask-RCNN model~\cite{he2017mask}. As shown in Fig.~\ref{fig:seg}, the top edge of the top textline and the bottom edge of the bottom textline in the PDF elements forms the top and bottom edge of the segmentation, respectively. The right edge of the textlines are scanned from top to bottom to form the right side of the segmentation. $\RADot$-shape edges are inserted if the right edge of a textline is on the left of the right edge of the textline above it. Otherwise, $\LADot$-shape edges are inserted. The left side of the segmentation is generated by scanning the left edge of the textlines from bottom to top using the same principle. For figure and table instances, the bounding box is reused as the segmentation, since almost all these instances are rectangular. Fig.~\ref{fig:annotation} illustrates the annotations for the PDF page in Fig.~\ref{fig:pdf_page}.

\subsubsection{Quality control}

There are several sources that can lead to discrepancies between the PDF parsing results and the corresponding XML. When discrepancies are over the threshold $d_{max}$, the annotation algorithm may not be able to identify all elements in a document page. For example, \texttt{PDFminer} parses some complex inline formulas completely differently from the XML, which leads to a large Levenshtein distance and failure to match PDF elements with XML nodes. Hence we need a way to evaluate how well a PDF page is annotated and eliminate poorly annotated pages from PubLayNet. The annotation quality of a PDF page is defined as the ratio of the area of textboxes, images, and geometric shapes that are annotated to the area of textboxes, images, and geometric shapes within the main text box of the page. Non-title pages of which the annotation quality is less than 99\% are excluded from PubLayNet, which is an extremely high standard to control the noise in PubLayNet at a low level. The format of title pages of different journals varies substantially. Miscellaneous information, such as manuscript history (dates of submission, revision, acceptance), copyright statement, editorial details, etc, is often included in title pages, but formatted differently from the XML representation and therefore missed in the annotations. To include adequate title pages, we set the threshold of annotation quality to 90\% for title pages.

\subsection{Data partition}

The annotated PDF pages are partitioned into training, development, and testing sets at journal level to maximize the differences between the sets. This allows better evaluation of how well a model generalizes to unseen paper templates.

The journals that contain $\le 2000$ pages, $\ge 320$ figures, $\ge 140$ tables, and $\ge 20$ lists are extracted to generate the development and testing sets. This avoids the development and testing sets from being dominated by a particular journal with a large number of pages, and ensures the development and testing sets have an adequate number instances of figures, tables, and lists.

Half of these journals are randomly drawn to generate the development set. The development set consists of all pages with a list in these journals, as well as 2000 title pages, 3000 pages with a table, 3000 pages with a figure, and 2000 plain pages, which are randomly drawn from these journals. The testing set is generated using the same procedure on the rest half of the journals. To further reduce the noise in the development and testing sets and make more valid evaluation of models, the development and testing sets are curated by human, where profound erroneous pages are removed and moderate erroneous pages are corrected.

The journals that do not satisfy the criteria above are used to generate the training set. To ensure diversity of the training data, from each of the journals, we randomly drawn at most 200 pages with a list, 50 pages with a table, 50 pages with a figure, 50 title pages, and 25 plain pages. The statistics of the training, development, and testing sets are depicted in detail in Table~\ref{tab:partition}.
\begin{table}[!htb]
  \caption{Statistics of training, development, and testing sets in PubLayNet. PubLayNet is one to two orders of magnitude larger than any existing document layout dataset.}
  \label{tab:partition}
  \centering
  \begin{threeparttable}
  \begin{tabular}{p{3cm}rrr}
    \toprule
    & Training & Development & Testing \\
    \midrule
    \textbf{Pages} & & & \\
    Plain pages & 87,608 & 1,138 & 1,121 \\
    Title pages & 46,480 & 2,059\tnote{+} & 2,021\tnote{+} \\
    Pages with lists & 53,793 & 2,984 & 3,207 \\
    Pages with tables & 86,950 & 3,772\tnote{+} & 3,950\tnote{+} \\
    Pages with figures & 96,656 & 3,734\tnote{+} & 3,807\tnote{+} \\
    Total & 340,391 & 11,858 & 11,983 \\
    \midrule
    \textbf{Instances} & & & \\
    Text & 2,376,702 & 93,528 & 95,780 \\
    Title & 633,359 & 19,908 & 20,340\\
    Lists & 81,850 & 4,561 & 5,156 \\
    Tables & 103,057 & 4,905 & 5,166 \\
    Figures & 116,692 & 4,913 & 5,333 \\
    Total & 3,311,660 & 127,815 & 131,775\\
    \bottomrule
\end{tabular}
\begin{tablenotes}\footnotesize
\item[+] These numbers are slightly greater than the number of pages drawn, as pages with lists may be title pages or contain tables or figures.
\end{tablenotes}
\end{threeparttable}
\end{table}

\section{Results}

Three experiments are designed to investigate 1) how well the established object detection models Faster-RCNN (F-RCNN)~\cite{ren2015faster} and Mask-RCNN (M-RCNN)~\cite{he2017mask} can recognize document layout of PubLayNet; 2) if the F-RCNN and M-RCNN models pre-trained on PubLayNet can be fine-tuned to tackle the ICDAR 2013 Table Recognition Competition\footnote{Other ICDAR competitions require annotations of document layout categories not available in PubLayNet and the size of the development sets is too small for effective training.}; 3) if the F-RCNN and M-RCNN models pre-trained on PubLayNet are better initializations than those pre-trained on the ImageNet and COCO datasets for analyzing documents in a different domain.

\subsection{Document layout recognition using deep learning}
\label{sec:exp_1}
We trained a F-RCNN model and a M-RCNN model on PubLayNet using the Detectron implementation~\cite{Detectron2018}. PDF pages are converted to images using the \texttt{pdf2image} package\footnote{\url{https://github.com/Belval/pdf2image}}. Each model was trained for 180k iterations with a base learning rate of 0.01. The learning rate was reduced by a factor of 10 at the 120k iteration and the 160k iteration. The models were trained on 8 GPUs with one image per GPU, which yields an effective mini-batch size of 8. Both models use the ResNeXt-101-64x4d model as the backbone, which was initialized with the model pre-trained on ImageNet.

The performance of the F-RCNN and the M-RCNN models on our development and testing sets are depicted in Table~\ref{tab:eval}. The evaluation metric is the mean average precision (MAP) @ intersection over union (IOU) [0.50:0.95] of bounding boxes, which is used in the COCO competition\footnote{\url{http://cocodataset.org/#detection-eval}}. Both models can generate accurate (MAP $> 0.9$) document layout, where M-RCNN shows a small advantage over F-RCNN. The models are more accurate at detecting tables and figures than texts, titles, and lists. We think this is attributed to more regular shapes, more distinctive differences from other categories, and lower rate of erroneous annotations in the training set. The models perform worst on titles, as titles are usually much smaller than other categories and more difficult to detect.
\begin{table}[!htb]
  \caption{MAP @ IOU [0.50:0.95] of the F-RCNN and the M-RCNN models on our development and testing sets. M-RCNN shows a small advantage over F-RCNN.}
  \label{tab:eval}
  \centering
  \begin{tabular}{lcccc}
    \toprule
    \multirow{2}{*}{Category} & \multicolumn{2}{c}{Dev} & \multicolumn{2}{c}{Test} \\
    & F-RCNN & M-RCNN & F-RCNN & M-RCNN \\
    \midrule
    Text & 0.910 & 0.916 & 0.913 & 0.917 \\
    Title & 0.826 & 0.840 & 0.812 & 0.828 \\
    List & 0.883 & 0.886 & 0.885 & 0.887 \\
    Table & 0.954 & 0.960 & 0.943 & 0.947 \\
    Figure & 0.937 & 0.949 & 0.945 & 0.955 \\
    \midrule
    Macro average & 0.902 & 0.910 & 0.900 & 0.907 \\
    \bottomrule
\end{tabular}
\end{table}

\begin{figure*}[!htb]
  \centering
  \includegraphics[width=\linewidth]{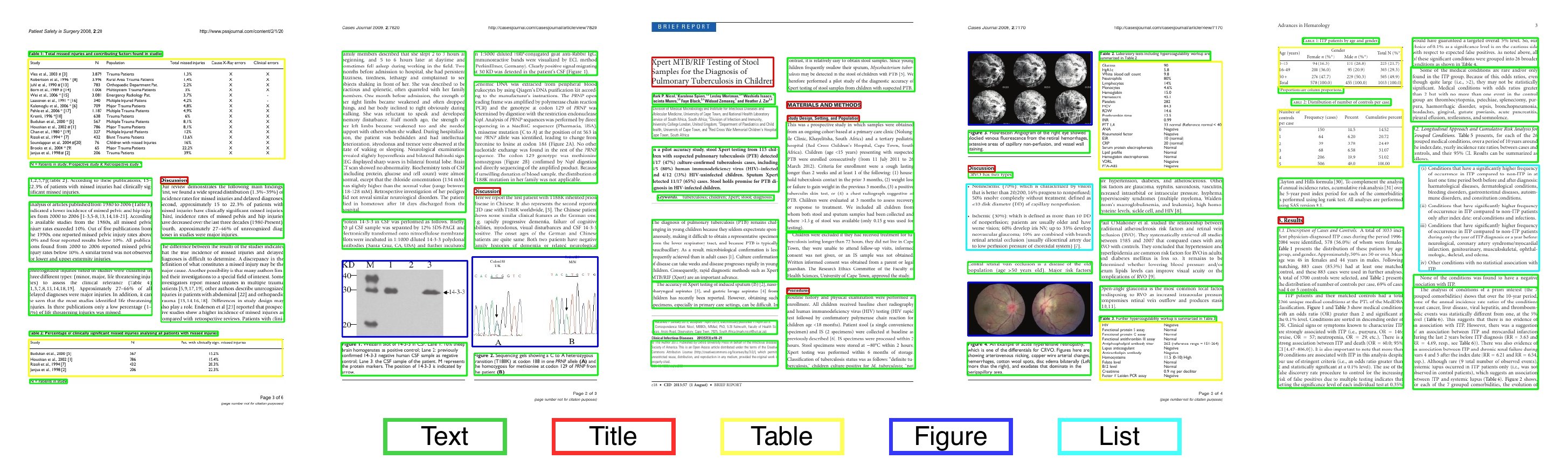}
  \caption{Representative examples of the document layout analysis results using the M-RCNN model. As implied by the high MAP, the model is able to generate accurate layout.}
  \label{fig:correct}
\end{figure*}

\begin{figure*}[!htb]
  \centering
  \includegraphics[width=\linewidth]{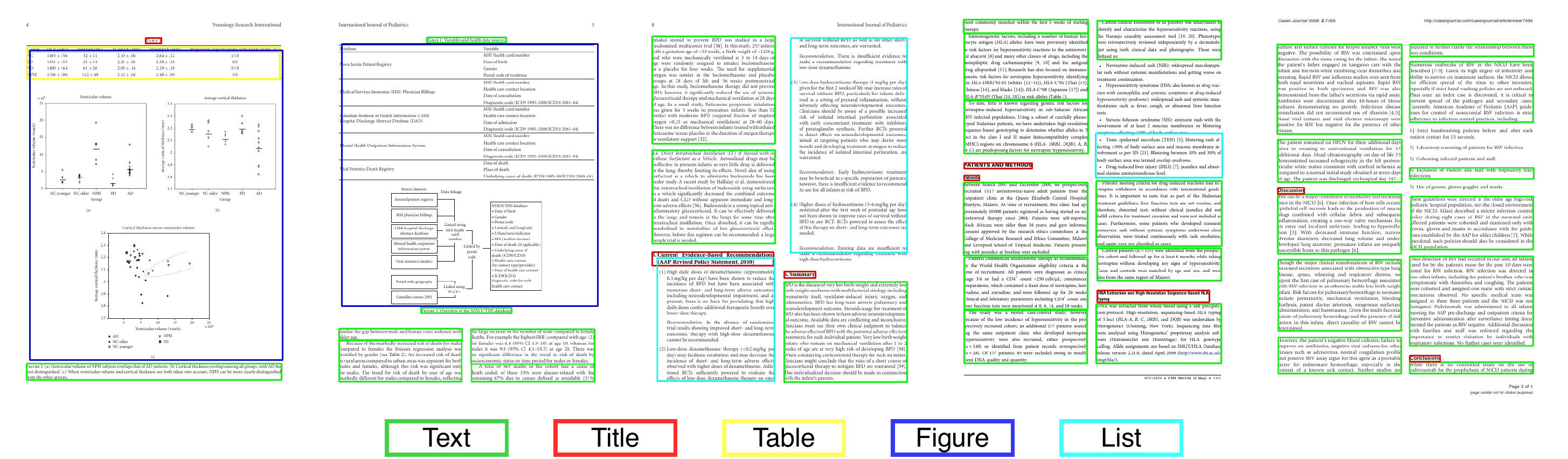}
  \caption{Erroneous document layout predictions made by the M-RCNN model.}
  \label{fig:errors}
\end{figure*}

Fig.~\ref{fig:correct} shows some representative examples of the document layout analysis results of testing pages using the M-RCNN model. As implied by the high MAP, the model is able to generate accurate layout. Fig.~\ref{fig:errors} illustrates some of the rare errors made by the M-RCNN model. We think some of the errors are attributed to the noise in PubLayNet. We will continue improving the quality of PubLayNet.

\subsection{Table detection}

The ICDAR 2013 Table Competition~\cite{gobel2013icdar} is one of the most prestigious competitions on table detection in PDF documents from European government sources. We created a table detection dataset by extracting from our training set the PDF pages that contain one or more tables, and remove non-table instances from the annotations. We trained a F-RCNN model and a M-RCNN model on this table detection dataset under the same configuration described in Section~\ref{sec:exp_1}. Then the models are fine-tuned with the 170 example PDF pages provided by the competition. For fine-tuning, we used a base learning rate 0.001, which was decreased by 10 at the 100th iteration out of 200 total iterations. The minimum confidence score for a detection is decided by a 5-fold cross-validation on the 170 training pages. The fine-tuned model was evaluated on the formal competition dataset (238 pages) using the official evaluation toolkit~\footnote{\url{https://github.com/tamirhassan/dataset-tools}}. Table~\ref{tab:ICDAR} compares the performance of the fine-tuned models and published approaches. The fine tuned F-RCNN model achieves the state-of-the-art performance reported in~\cite{schreiber2017deepdesrt}, where the F-RCNN model was fine tuned with 1600 samples from a pre-trained object detection model. By fine tuning from a model pre-trained on document samples, we can obtain the same level of performance with much smaller training data (170 samples).
\begin{table}[!htb]
  \caption{Fine tuning pre-trained F-RCNN and M-RCNN models for the ICDAR 2013 Table Recognition Competition. Based on models pre-trained on table pages in PubLayNet, we obtained the state-of-the-art performance with only 170 training pages.}
  \label{tab:ICDAR}
  \centering
  \begin{tabular}{llccc}
    \toprule
    Input & Method & Precision & Recall & F1-score \\
    \midrule
    \multirow{4}{*}{Image} &
    \textbf{F-RCNN} & 0.972 & 0.964 & 0.968 \\
    & \textbf{M-RCNN} & 0.940 & 0.955 & 0.947 \\
    & Schreiber et al. 2017~\cite{schreiber2017deepdesrt} & 0.974 & 0.962 & 0.968 \\
    & Tran et al. 2015~\cite{tran2015table} & 0.952 & 0.964 & 0.958 \\
    \midrule
    \multirow{4}{*}{PDF} & Hao et al. 2016~\cite{hao2016table} & 0.972 & 0.922 & 0.946 \\
    & Silva 2010~\cite{silva2010parts} & 0.929 & 0.983 & 0.955 \\
    & Nurminen 2013~\cite{gobel2013icdar} & 0.921 & 0.908 & 0.914 \\
    & Yildiz 2005~\cite{yildiz2005pdf2table} & 0.640 & 0.853 & 0.731 \\
    \bottomrule
\end{tabular}
\end{table}

\subsection{Fine tuning for a different domain}

In USA, there is a large number of private health insurance providers. Employees are provided with Summary Plan Description (SPD) documents typically in PDF, which describe the benefits provided by the private health insurers. There is a large variety in the layout of SPD documents provided by different companies. The layout of these documents is also distinctively different from scientific publications.

We manually annotated the texts, tables, and lists in 20 representative SPD documents that cover a large number of possible layouts. This domain specific dataset contains 2,131 pages, 9,379, 2,500, and 820 instances of text, tables, and lists, respectively. A 5-fold cross-document-validation\footnote{For each fold, the model is trained on 16 documents and tested on 4 documents} was taken to compare different pre-trained F-RCNN and M-RCNN models for fine tuning.

We evaluated three fine tuning approaches: 1) initializing the backbone with pre-trained ImageNet model, 2) initializing the whole model with pre-trained COCO model, and 3) initializing the whole model with pre-trained PubLayNet model. We also tested the zero-shot performance of the pre-trained PubLayNet model. The comparison of the performance of the approaches is illustrated in Table~\ref{tab:fine_tuning}. The performance of the zero-shot PubLayNet model is considerably worse than the fine-tuned models, which demonstrates the distinct difference between the layout of SPD documents and PubMed Central\textsuperscript{\texttrademark} articles. Fine tuning the pre-trained PubLayNet model can substantially outperform other fine-tuned models. This demonstrates the advantage of using PubLayNet for document layout analysis. The only exception is that fine tuning pre-trained COCO F-RCNN model detects tables more accurately than fine tuning pre-trained PubLayNet F-RCNN model. In addition, the improvement on table detection by fine tuning pre-trained PubLayNet MRCNN model is relatively low to that on text and list detection. We think this is because the difference of table styles between SPD and the PubMed Central\textsuperscript{\texttrademark} articles and more substantial than that of text and list styles, and therefore less knowledge can be transferred to the fine tuned model.
\begin{table}[!htb]
  \caption{Comparison of the performance of fine tuning different pre-trained F-RCNN and M-RCNN models on the SPD documents. The performance scores are MAP @ IOU [0.50:0.95] evaluated via a 5-fold cross-document-validation on 20 SPD documents. The results demonstrate the advantage of PubLayNet over general image datasets in domain adaptation for document layout analysis.}
  \label{tab:fine_tuning}
  \centering
  \begin{tabular}{lllcc}
    \toprule
    Category & Method & Initialization & F-RCNN & M-RCNN \\
    \midrule
    \multirow{4}{*}{Text} & zero-shot & PubLayNet & 0.482 & 0.468 \\
    & fine tuning & PubLayNet & \textbf{0.701} & \textbf{0.708} \\
    & fine tuning & COCO & 0.651 & 0.661 \\
    & fine tuning & ImageNet & 0.622 & 0.629 \\
    \midrule
    \multirow{4}{*}{List} & zero-shot & PubLayNet & 0.508 & 0.510 \\
    & fine tuning & PubLayNet & \textbf{0.681} & \textbf{0.684} \\
    & fine tuning & COCO & 0.622 & 0.611 \\
    & fine tuning & ImageNet & 0.603 & 0.603 \\
    \midrule
    \multirow{4}{*}{Table} & zero-shot & PubLayNet & 0.422 & 0.419 \\
    & fine tuning & PubLayNet & 0.541 & \textbf{0.596} \\
    & fine tuning & COCO & \textbf{0.560} & 0.588 \\
    & fine tuning & ImageNet & 0.528 & 0.573 \\
    \midrule
    \multirow{4}{1cm}{Macro average} & zero-shot & PubLayNet & 0.470 & 0.465 \\
    & fine tuning & PubLayNet & \textbf{0.641} & \textbf{0.663} \\
    & fine tuning & COCO & 0.611 & 0.620 \\
    & fine tuning & ImageNet & 0.584 & 0.602 \\
    \bottomrule
\end{tabular}
\end{table}

\section{Discussion}

PMCOA provides a large set of documents available at the same time in PDF and XML format.
The methodology proposed in this work to generate a large dataset of article pages automatically annotated with the location of document layout components using the PMCOA documents.
The quality assurance has shown that the automatically generated annotations are of high quality.
In addition, existing state-of-the-art object detection algorithms reproduce successfully the annotations from the automatically annotated set.
The title category seems to be the weakest one.
The identification of titles is challenging due to the different ways in which titles are present in the documents.
On the other hand, titles are identified as text in ICDAR competitions and titles could be merged with the text category in this set up.

The documents in PubLayNet are all scientific literature, which is domain specific and limits the heterogeneity of layout. We took several methods in developing and partition PubLayNet to utilize as much variation in PMCOA as possible and prevent PubLayNet from being dominated by a certain journal. With over 6,500 journals included in PMCOA, our experiment shows that the training set is heterogeneous enough to train deep learning models that can accurately recognize the layout of unseen journals. For documents in a distant domain, e.g., government documents and SPD documents, we demonstrated the value of using PubLayNet in a transfer learning setting.

\section{Conclusion}

We automatically generated the PubLayNet dataset, which is the largest ever available document layout annotation dataset exploiting redundancy in PCMOA.
This dataset allows state-of-the-art object detection algorithms to be trained delivering high performance layout recognition on biomedical articles.
Furthermore, this dataset is shown to be helpful to pre-train object detection algorithms to identify tables and different document layout objects in health insurance documents.
These results are encouraging since the developed dataset is potentially helpful for document layout annotation of other domains.
PubLayNet is available from \url{https://github.com/ibm-aur-nlp/PubLayNet}.

As future work, we plan to exploit PMCOA for the automatic generation of large  datasets to solve other document analysis problems with deep learning models. For example, PubLayNet does not contain relationships between the layout elements, e.g., a paragraph and a section title. Such information is available in the XML representation and can be exploit to automatically create a dataset of the logical structure of documents.

\section*{Acknowledgment}
We thank Manoj Gambhir for relevant feedback on this work and for his work in the development of the SPD data set and Shaila Pervin for her work in the development of the SPD data set.

\bibliographystyle{IEEEtran}
\bibliography{kdd-2019}

\begin{thebibliography}{10}
\providecommand{\url}[1]{#1}
\csname url@samestyle\endcsname
\providecommand{\newblock}{\relax}
\providecommand{\bibinfo}[2]{#2}
\providecommand{\BIBentrySTDinterwordspacing}{\spaceskip=0pt\relax}
\providecommand{\BIBentryALTinterwordstretchfactor}{4}
\providecommand{\BIBentryALTinterwordspacing}{\spaceskip=\fontdimen2\font plus
\BIBentryALTinterwordstretchfactor\fontdimen3\font minus
  \fontdimen4\font\relax}
\providecommand{\BIBforeignlanguage}[2]{{%
\expandafter\ifx\csname l@#1\endcsname\relax
\typeout{** WARNING: IEEEtran.bst: No hyphenation pattern has been}%
\typeout{** loaded for the language `#1'. Using the pattern for}%
\typeout{** the default language instead.}%
\else
\language=\csname l@#1\endcsname
\fi
#2}}
\providecommand{\BIBdecl}{\relax}
\BIBdecl

\bibitem{staar2018ccs}
\BIBentryALTinterwordspacing
P.~W.~J. Staar, M.~Dolfi, C.~Auer, and C.~Bekas, ``Corpus conversion service: A
  machine learning platform to ingest documents at scale,'' in
  \emph{Proceedings of the 24th ACM SIGKDD International Conference on
  Knowledge Discovery \&\#38; Data Mining}, ser. KDD '18.\hskip 1em plus 0.5em
  minus 0.4em\relax New York, NY, USA: ACM, 2018, pp. 774--782. [Online].
  Available: \url{http://doi.acm.org/10.1145/3219819.3219834}
\BIBentrySTDinterwordspacing

\bibitem{cattoni1998geometric}
R.~Cattoni, T.~Coianiz, S.~Messelodi, and C.~M. Modena, ``Geometric layout
  analysis techniques for document image understanding: a review,''
  \emph{ITC-irst Technical Report}, vol. 9703, no.~09, 1998.

\bibitem{breuel2002two}
T.~M. Breuel, ``Two geometric algorithms for layout analysis,'' in
  \emph{International workshop on document analysis systems}.\hskip 1em plus
  0.5em minus 0.4em\relax Springer, 2002, pp. 188--199.

\bibitem{breuel2003high}
------, ``High performance document layout analysis,'' in \emph{Proceedings of
  the Symposium on Document Image Understanding Technology}, 2003, pp.
  209--218.

\bibitem{he2017mask}
K.~He, G.~Gkioxari, P.~Doll{\'a}r, and R.~Girshick, ``Mask r-cnn,'' in
  \emph{Computer Vision (ICCV), 2017 IEEE International Conference on}.\hskip
  1em plus 0.5em minus 0.4em\relax IEEE, 2017, pp. 2980--2988.

\bibitem{schreiber2017deepdesrt}
S.~Schreiber, S.~Agne, I.~Wolf, A.~Dengel, and S.~Ahmed, ``Deepdesrt: Deep
  learning for detection and structure recognition of tables in document
  images,'' in \emph{Document Analysis and Recognition (ICDAR), 2017 14th IAPR
  International Conference on}, vol.~1.\hskip 1em plus 0.5em minus 0.4em\relax
  IEEE, 2017, pp. 1162--1167.

\bibitem{antonacopoulos2009realistic}
A.~Antonacopoulos, D.~Bridson, C.~Papadopoulos, and S.~Pletschacher, ``A
  realistic dataset for performance evaluation of document layout analysis,''
  in \emph{Document Analysis and Recognition, 2009. ICDAR'09. 10th
  International Conference on}.\hskip 1em plus 0.5em minus 0.4em\relax IEEE,
  2009, pp. 296--300.

\bibitem{clausner2015enp}
C.~Clausner, C.~Papadopoulos, S.~Pletschacher, and A.~Antonacopoulos, ``The enp
  image and ground truth dataset of historical newspapers,'' in \emph{Document
  Analysis and Recognition (ICDAR), 2015 13th International Conference
  on}.\hskip 1em plus 0.5em minus 0.4em\relax IEEE, 2015, pp. 931--935.

\bibitem{clausner2017icdar2017}
C.~Clausner, A.~Antonacopoulos, and S.~Pletschacher, ``Icdar2017 competition on
  recognition of documents with complex layouts-rdcl2017,'' in \emph{Document
  Analysis and Recognition (ICDAR), 2017 14th IAPR International Conference
  on}, vol.~1.\hskip 1em plus 0.5em minus 0.4em\relax IEEE, 2017, pp.
  1404--1410.

\bibitem{shahab2010open}
A.~Shahab, F.~Shafait, T.~Kieninger, and A.~Dengel, ``An open approach towards
  the benchmarking of table structure recognition systems,'' in
  \emph{Proceedings of the 9th IAPR International Workshop on Document Analysis
  Systems}.\hskip 1em plus 0.5em minus 0.4em\relax ACM, 2010, pp. 113--120.

\bibitem{fang2012dataset}
J.~Fang, X.~Tao, Z.~Tang, R.~Qiu, and Y.~Liu, ``Dataset, ground-truth and
  performance metrics for table detection evaluation,'' in \emph{Document
  Analysis Systems (DAS), 2012 10th IAPR International Workshop on}.\hskip 1em
  plus 0.5em minus 0.4em\relax IEEE, 2012, pp. 445--449.

\bibitem{yildiz2005pdf2table}
B.~Yildiz, K.~Kaiser, and S.~Miksch, ``pdf2table: A method to extract table
  information from pdf files,'' in \emph{IICAI}, 2005, pp. 1773--1785.

\bibitem{tran2015table}
D.~N. Tran, T.~A. Tran, A.~Oh, S.~H. Kim, and I.~S. Na, ``Table detection from
  document image using vertical arrangement of text blocks,''
  \emph{International Journal of Contents}, vol.~11, no.~4, pp. 77--85, 2015.

\bibitem{he2017multi}
D.~He, S.~Cohen, B.~Price, D.~Kifer, and C.~L. Giles, ``Multi-scale multi-task
  fcn for semantic page segmentation and table detection,'' in \emph{Document
  Analysis and Recognition (ICDAR), 2017 14th IAPR International Conference
  on}, vol.~1.\hskip 1em plus 0.5em minus 0.4em\relax IEEE, 2017, pp. 254--261.

\bibitem{hao2016table}
L.~Hao, L.~Gao, X.~Yi, and Z.~Tang, ``A table detection method for pdf
  documents based on convolutional neural networks,'' in \emph{Document
  Analysis Systems (DAS), 2016 12th IAPR Workshop on}.\hskip 1em plus 0.5em
  minus 0.4em\relax IEEE, 2016, pp. 287--292.

\bibitem{gilani2017table}
A.~Gilani, S.~R. Qasim, I.~Malik, and F.~Shafait, ``Table detection using deep
  learning,'' in \emph{2017 14th IAPR International Conference on Document
  Analysis and Recognition (ICDAR)}, vol.~01, Nov 2017, pp. 771--776.

\bibitem{kavasidis2018saliency}
I.~Kavasidis, S.~Palazzo, C.~Spampinato, C.~Pino, D.~Giordano, D.~Giuffrida,
  and P.~Messina, ``A saliency-based convolutional neural network for table and
  chart detection in digitized documents,'' \emph{arXiv preprint
  arXiv:1804.06236}, 2018.

\bibitem{levenshtein1966binary}
V.~I. Levenshtein, ``Binary codes capable of correcting deletions, insertions,
  and reversals,'' in \emph{Soviet physics doklady}, vol.~10, no.~8, 1966, pp.
  707--710.

\bibitem{ren2015faster}
S.~Ren, K.~He, R.~Girshick, and J.~Sun, ``Faster r-cnn: Towards real-time
  object detection with region proposal networks,'' in \emph{Advances in neural
  information processing systems}, 2015, pp. 91--99.

\bibitem{Detectron2018}
R.~Girshick, I.~Radosavovic, G.~Gkioxari, P.~Doll\'{a}r, and K.~He,
  ``Detectron,'' \url{https://github.com/facebookresearch/detectron}, 2018.

\bibitem{gobel2013icdar}
M.~G{\"o}bel, T.~Hassan, E.~Oro, and G.~Orsi, ``Icdar 2013 table competition,''
  in \emph{Document Analysis and Recognition (ICDAR), 2013 12th International
  Conference on}.\hskip 1em plus 0.5em minus 0.4em\relax IEEE, 2013, pp.
  1449--1453.

\bibitem{silva2010parts}
A.~Silva, ``Parts that add up to a whole: a framework for the analysis of
  tables,'' \emph{Edinburgh University, UK}, 2010.

\end{thebibliography}

\end{document}